\documentclass[10pt,twocolumn,letterpaper]{article}

\usepackage{cvpr}
\usepackage{times}
\usepackage{epsfig}
\usepackage{graphicx}
\usepackage{amsmath}
\usepackage{amssymb}

\usepackage[pagebackref=true,breaklinks=true,letterpaper=true,colorlinks,bookmarks=false]{hyperref}

\usepackage{amsmath}
\usepackage{bbold}
\usepackage{booktabs}
\usepackage{diagbox}
\usepackage{graphbox}
\usepackage{float}
\usepackage{footnote}
\usepackage{makecell}
\usepackage{multirow}
\usepackage{rotating}
\usepackage{setspace}
\usepackage{sidecap}
\usepackage{tabularx}
\usepackage{mathrsfs}
\usepackage{subcaption}
\usepackage{wrapfig}
\usepackage[capitalise]{cleveref}
\usepackage{xspace}
\usepackage{dsfont}
\makeatletter
\DeclareRobustCommand\onedot{\futurelet\@let@token\@onedot}
\def\@onedot{\ifx\@let@token.\else.\null\fi\xspace}

\def\etal{\emph{et al}\onedot}
\makeatother

\newcolumntype{L}{>{\raggedright\arraybackslash}X}
\newcolumntype{C}{>{\centering\arraybackslash}X}

\DeclareMathOperator*{\argmax}{argmax}

\newcommand{\norm}[1]{\left\lVert#1\right\rVert}

\usepackage[dvipsnames]{xcolor}
\newcommand{\rev}[1]{#1}

\cvprfinalcopy 

\def\cvprPaperID{2452} 

\begin{document}

\title{Supervised Fitting of Geometric Primitives to 3D Point Clouds}

\author{Lingxiao Li*\textsuperscript{1} $\quad$
Minhyuk Sung*\textsuperscript{1} $\quad$
Anastasia Dubrovina\textsuperscript{1} $\quad$
Li Yi\textsuperscript{1} $\quad$
Leonidas Guibas\textsuperscript{1,2} \\
\textsuperscript{1}Stanford University $\quad$ \textsuperscript{2}Facebook AI Research\\
}

\maketitle
\setlength\abovedisplayskip{5pt}
\setlength\belowdisplayskip{5pt}

\newcommand\blfootnote[1]{%
  \begingroup
  \renewcommand\thefootnote{}\footnote{#1}%
  \addtocounter{footnote}{-1}%
  \endgroup
}
\blfootnote{*equal contribution}


\begin{abstract}
\vspace{-5pt}
Fitting geometric primitives to 3D point cloud data bridges a gap between low-level digitized 3D data and high-level structural information on the underlying 3D shapes. As such, it enables many downstream applications in 3D data processing. For a long time, RANSAC-based methods have been the gold standard for such primitive fitting problems, but they require careful per-input parameter tuning and thus do not scale well for large datasets with diverse shapes. In this work, we introduce Supervised Primitive Fitting Network (SPFN), an end-to-end neural network that can robustly detect a varying number of primitives at different scales without any user control. The network is supervised using ground truth primitive surfaces and primitive membership for the input points. Instead of directly predicting the primitives, our architecture first predicts per-point properties and then uses a differential model estimation module to compute the primitive type and parameters. We evaluate our approach on a novel benchmark of ANSI 3D mechanical component models and demonstrate a significant improvement over both the state-of-the-art RANSAC-based methods and the direct neural prediction.
\end{abstract}
\vspace{-\baselineskip}


\section{Introduction}
\label{sec:introduction}
\begin{figure}[t]
  \centering
  \includegraphics[width=\columnwidth]{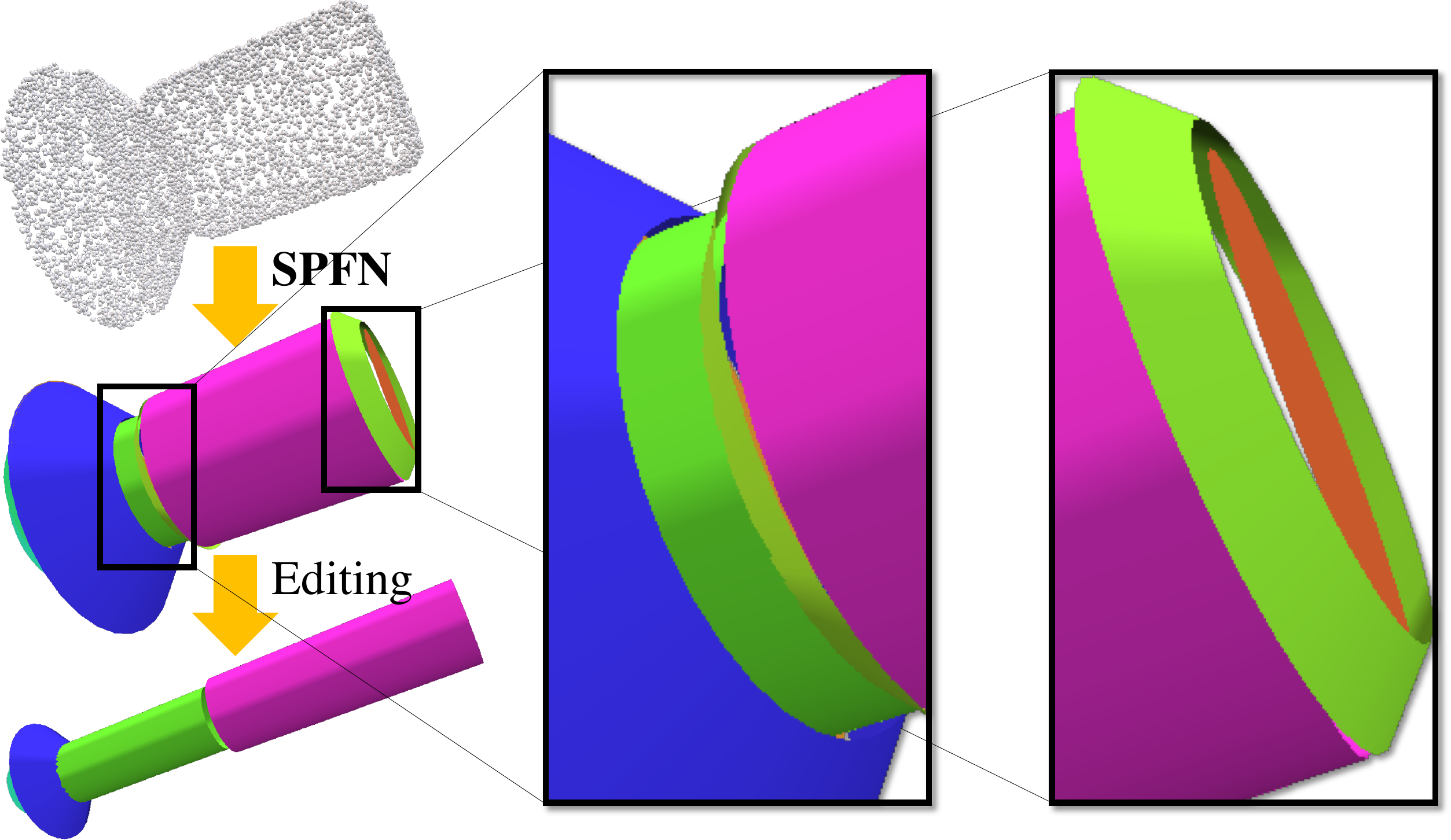}
  \caption{Our network SPFN generates a collection of geometric primitives that fit precisely to the input point cloud, even for tiny segments. The predicted primitives can then be used for structure understanding or shape editing.}
  \label{fig:teaser}
\vspace{-1.5\baselineskip}
\end{figure}

\vspace{-3pt}
Recent 3D scanning techniques and large-scale 3D repositories have widened opportunities for 3D geometric data processing.
However, most of the scanned data and the models in these repositories are represented as digitized point clouds or meshes. Such low-level representations of 3D data limit our ability to geometrically manipulate them due to the lack of structural information aligned with the shape semantics.
For example, when editing a shape built from geometric primitives, the knowledge of the type and parameters of each primitive can greatly aid the manipulation in producing a plausible result (Figure~\ref{fig:teaser}). 
To address the absence of such structural information in digitized data, in this work we consider the conversion problem of mapping a 3D point cloud to a number of geometric primitives that best fit the underlying shape.

Representing an object with a set of simple geometric components is a long-standing problem in computer vision. Since 
the 1970s~\cite{Binford:1971,Marr:1978}, the fundamental ideas for tackling the problem have been revised by many researchers, even until recently~\cite{Tulsiani:2017,Zou:2017,Ganapathi:2018}.
However, most of these previous work aimed at solving \emph{perceptual} learning tasks; the main focus was on parsing shapes, or generating a rough abstraction of the geometry with bounding primitives. In contrast, our goal is set at precisely fitting geometric primitives to the shape \emph{surface}, even with the presence of noise in the input.

For this primitive fitting problem, RANSAC-based methods~\cite{Schnabel:2007} remain the standard.
The main drawback of these approaches is the difficulty of finding suitable algorithm parameters. 
For example, if the threshold of fitting residual for accepting a candidate primitive is smaller than the noise level, over-segmentation may occur, whereas a too large threshold will cause the algorithm to miss small pieces primitives. 
This problem happens not only when processing noisy scanned data, but also when parsing meshes in 3D repositories because the discretization of the original shape into the mesh obscures the accurate local geometry of the shape surface. 
The demand for careful user control prevents RANSAC-based methods to scale up to a large number of categories of diverse shapes.


Such drawback motivates us to consider a \emph{supervised} deep learning framework.
The primitive fitting problem can be viewed as a model prediction problem, and the simplest approach would be directly regressing the parameters in the parameter space using a neural network. 
However, the regression loss based on direct measurement of the parameter difference does not reflect the actual fitting error -- the distances between input points and the primitives. 
Such misinformed loss function can significantly limit prediction accuracy. 
To overcome this, Brachmann \etal~\cite{Brachmann:2017} integrated the RANSAC pipeline into an end-to-end neural network by replacing the hypothesis selection step with a differentiable procedure.
However, their framework predicts only a single model, and it is not straightforward to extend it to predict \emph{multiple} models (primitives in our case).
Ranftl \etal~\cite{Ranftl:2018} also introduced a deep learning framework to perform model fitting via inlier weight prediction. We extend this idea to predict weights representing per-point membership for \emph{multiple} primitive models in our setting.

In this work, we propose Supervised Primitive Fitting Network (SPFN) that takes point clouds as input and predicts a varying number of primitives of different types with accurate parameters. 
For robust estimation, SPFN does not directly output primitive parameters, but instead predicts three kinds of per-point properties: point-to-primitive membership, surface normal, and the type of the primitive the point belongs to. Our framework supports four types of primitives: plane, sphere, cylinder, and cones. These types form the most major components in CAD models. 
Given these per-point properties, our \emph{differentiable} model estimator computes the primitive parameters in an algebraic way, making the fitting loss fully backpropable.
The advantage of our approach is that the network can leverage the readily available supervisions of per-point properties in training. 
It has been shown that per-point classification problems (membership, type) are suitable to address using a neural network that directly consumes a point cloud as input~\cite{Qi:2017a, Qi:2017b}.
Normal prediction can also be handled effectively with a similar neural network~\cite{Bansal:2016,Guerrero:2018}.

We train and evaluate the proposed method using our novel dataset, ANSI 3D mechanical component models with $17k$ CAD models.
The supervision in training is provided by parsing the CAD models and extracting the primitive information.
In our comparison experiments, we demonstrate that our supervised approach outperforms the widely used RANSAC-based approach~\cite{Schnabel:2007} with a big margin, despite using models from \emph{separate} categories in training and testing. Our method shows better fitting accuracy compared to~\cite{Schnabel:2007} even when we provide the latter with much higher-resolution point clouds as input.

\paragraph{Key contributions}
\vspace{-1\baselineskip}
\begin{itemize}
  \setlength\itemsep{-0.2em}
  \item
  We propose SPFN, an end-to-end supervised neural network that takes a point cloud as input and detects a varying number of primitives with different scales.
  \item 
  Our differentiable primitive model estimator solves a series of linear least-square problems, thus making the whole pipeline end-to-end trainable.
  \item We demonstrate the performance of our network using a novel CAD model dataset of mechanical components.
\end{itemize}


\section{Related Work}
\label{sec:related_work}
\vspace{-3pt}
\rev{Among a large body of previous work on fitting primitives to 3D data, we review only methods that fit primitives to \emph{objects} instead of scenes, as our target use cases are scanned point clouds of individual mechanical parts. For a more comprehensive review, see survey~\cite{Kaiser:2018}}.

\vspace{-\baselineskip}
\paragraph{RANSAC-based Primitive Fitting.}
RANSAC~\cite{Fischler:1981} and its variants~\cite{Torr:2000,Matas:2004,Chum:2005,Kang:2015} are the most widely used methods for primitive detection
in computer vision. 
A significant recent paper by Schnabel \etal~\cite{Schnabel:2007} introduced a robust RANSAC-based framework for detecting multiple primitives of different types in a dense point cloud. Li \etal~\cite{Li:2011} extended \cite{Schnabel:2007} by introducing a follow-up optimization that refines the extracted primitives based on the relations among them. 
As a downstream application of the RANSAC-based methods, Wu \etal~\cite{Wu:2018} and Du \etal~\cite{Du:2018} proposed a procedure to reverse-engineer the Constructive Solid Geometry (CSG) model from an input point cloud or mesh. While these RANSAC variants showed state-of-the-art results in their respective fields, their performance typically depends on careful and laborious parameter tuning for each category of shapes. In addition, point normals are required, which are not readily available from 3D scans. 
In contrast, our supervised deep learning architecture requires only point cloud data as input and does not need any user control at test time.

\vspace{-\baselineskip}
\paragraph{Network-based Primitive Fitting.}
Neural networks have been used in recent approaches to solve the primitive fitting problem in both supervised~\cite{Zou:2017} and unsupervised~\cite{Tulsiani:2017, Sharma:2018} settings. However, these methods are limited in accuracy with a restricted number of supported types. 
In the work of Zou \etal~\cite{Zou:2017} and Tulsiani \etal~\cite{Tulsiani:2017}, only cuboids are predicted and therefore can only serve as a rough abstraction of the input shape or image.
CSGNet~\cite{Sharma:2018} is capable of predicting more variety of primitives but with low accuracy, as the parameter extraction is done by performing \emph{classification} on a discretized parameter space. In addition, their reinforcement learning step requires rendering a CSG model to generate visual feedback for every training iteration, making the computation demanding. 
Our framework can be trained end-to-end and thus does not need expensive external procedures.

\section{Supervised Primitive Fitting Network}
\label{sec:framework}
\vspace{-3pt}

\begin{figure*}[t!]
  \includegraphics[width=\textwidth]{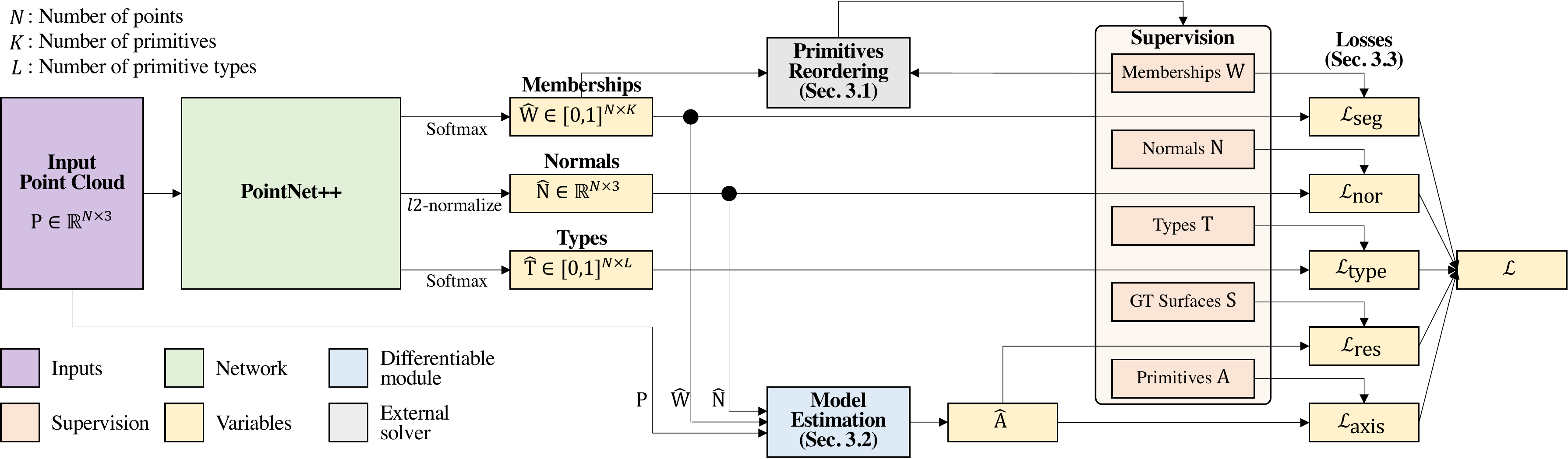}
\caption{Network architecture. PointNet++~\cite{Qi:2017b} takes input point cloud $\mathbf{P}$ and outputs three per-point properties: point-to-primitive membership $\mathbf{\hat{W}}$, normals $\mathbf{\hat{N}}$, and associated primitive type $\mathbf{\hat{T}}$. The order of ground truth primitives are matched with the output in the primitive reordering step (Section~\ref{sec:primitives_reordering}). Then, the output primitive parameters are estimated from the point properties in the model estimations step (Section~\ref{sec:model_estimation}). The loss is defined as the sum of five loss terms  (Section~\ref{sec:loss_function}).}
\label{fig:pipeline}
\vspace{-\baselineskip}
\end{figure*}

\begin{figure}[t!]
\centering
\begin{minipage}[t]{.25\columnwidth}
  \includegraphics[width=\linewidth,trim={0.8cm 0.8cm 0.8cm 0.8cm},clip]{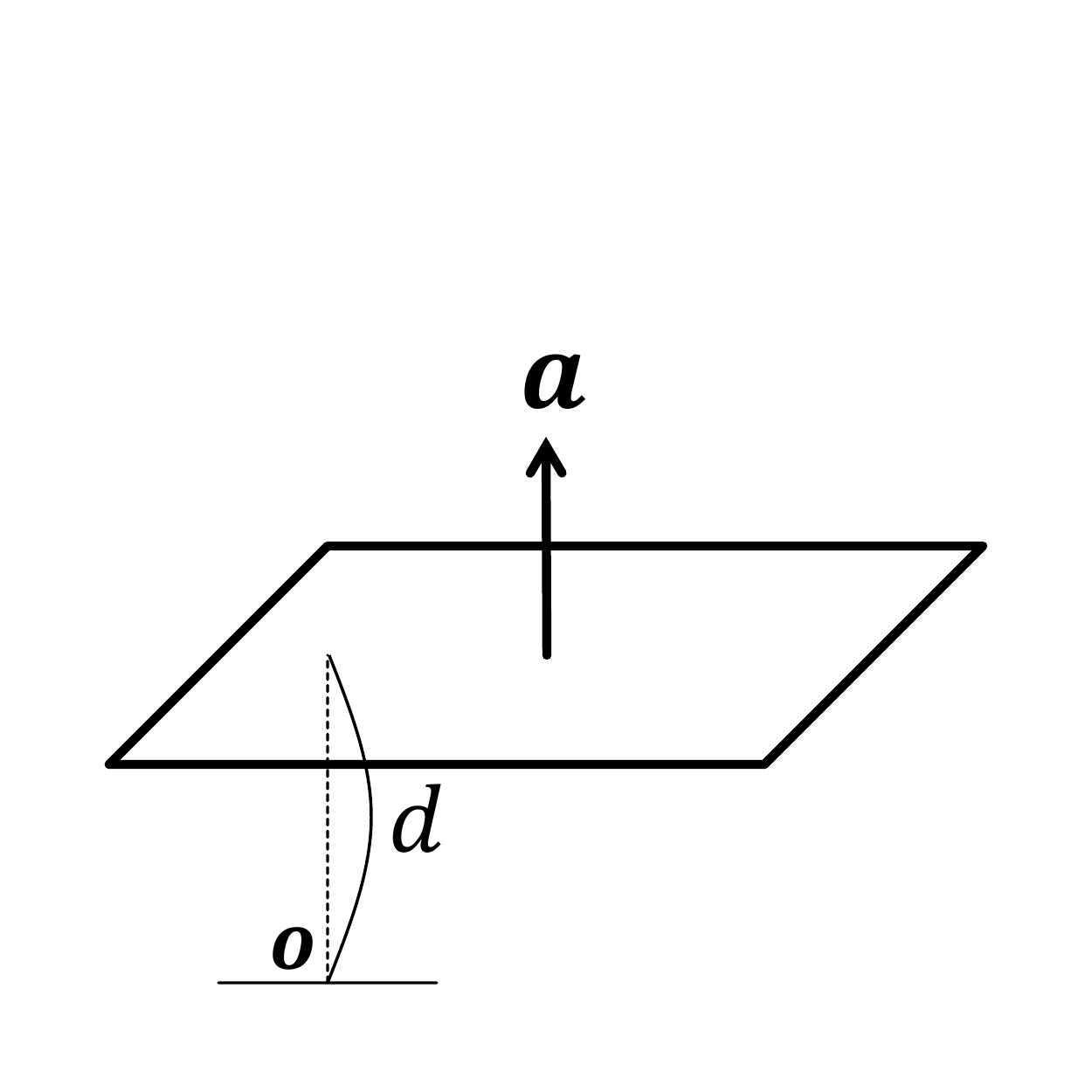}
\end{minipage}%
\begin{minipage}[t]{.25\columnwidth}
  \includegraphics[width=\linewidth,trim={0.8cm 0.8cm 0.8cm 0.8cm},clip]{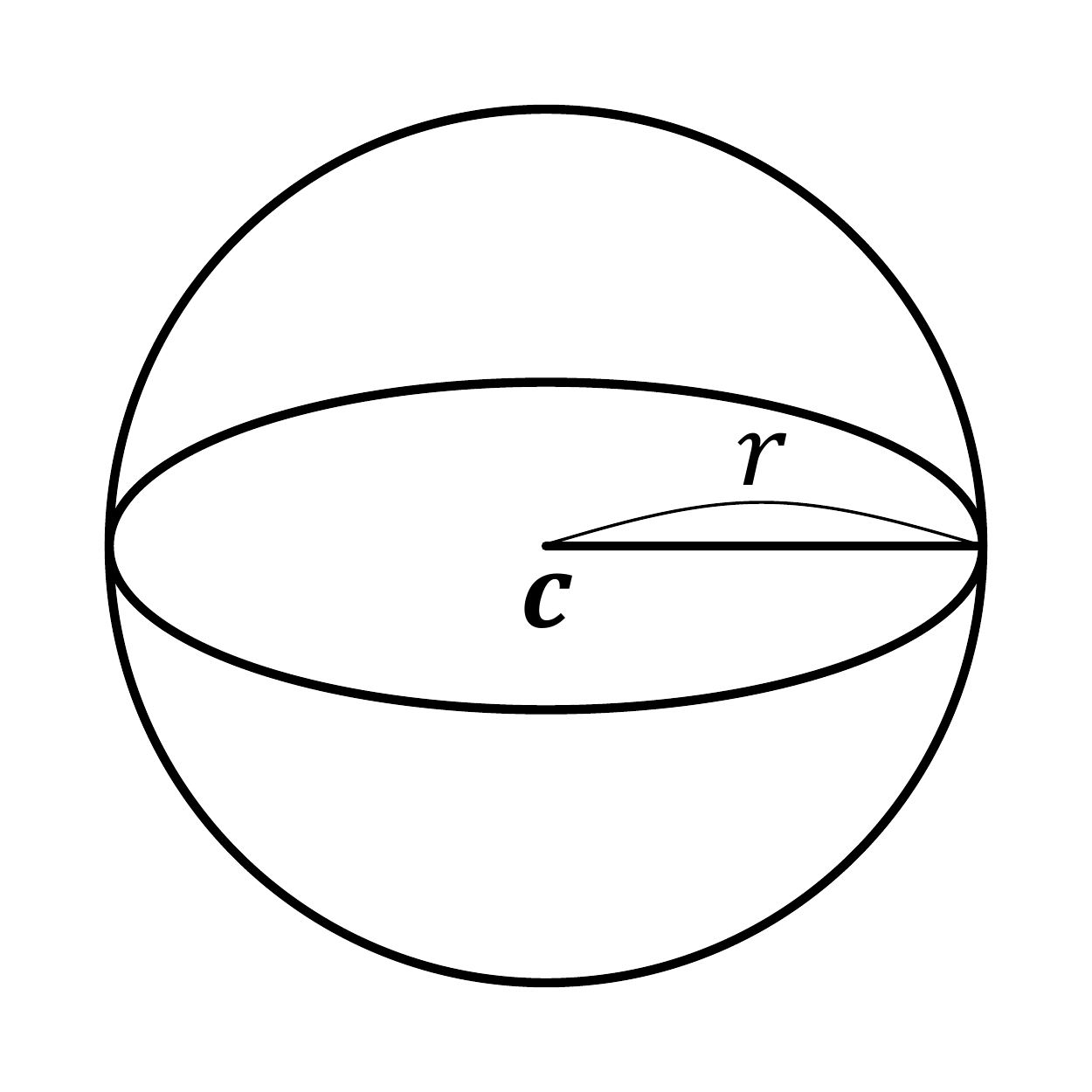}
\end{minipage}%
\begin{minipage}[t]{.25\columnwidth}
  \includegraphics[width=\linewidth,trim={0.8cm 0.8cm 0.8cm 0.8cm},clip]{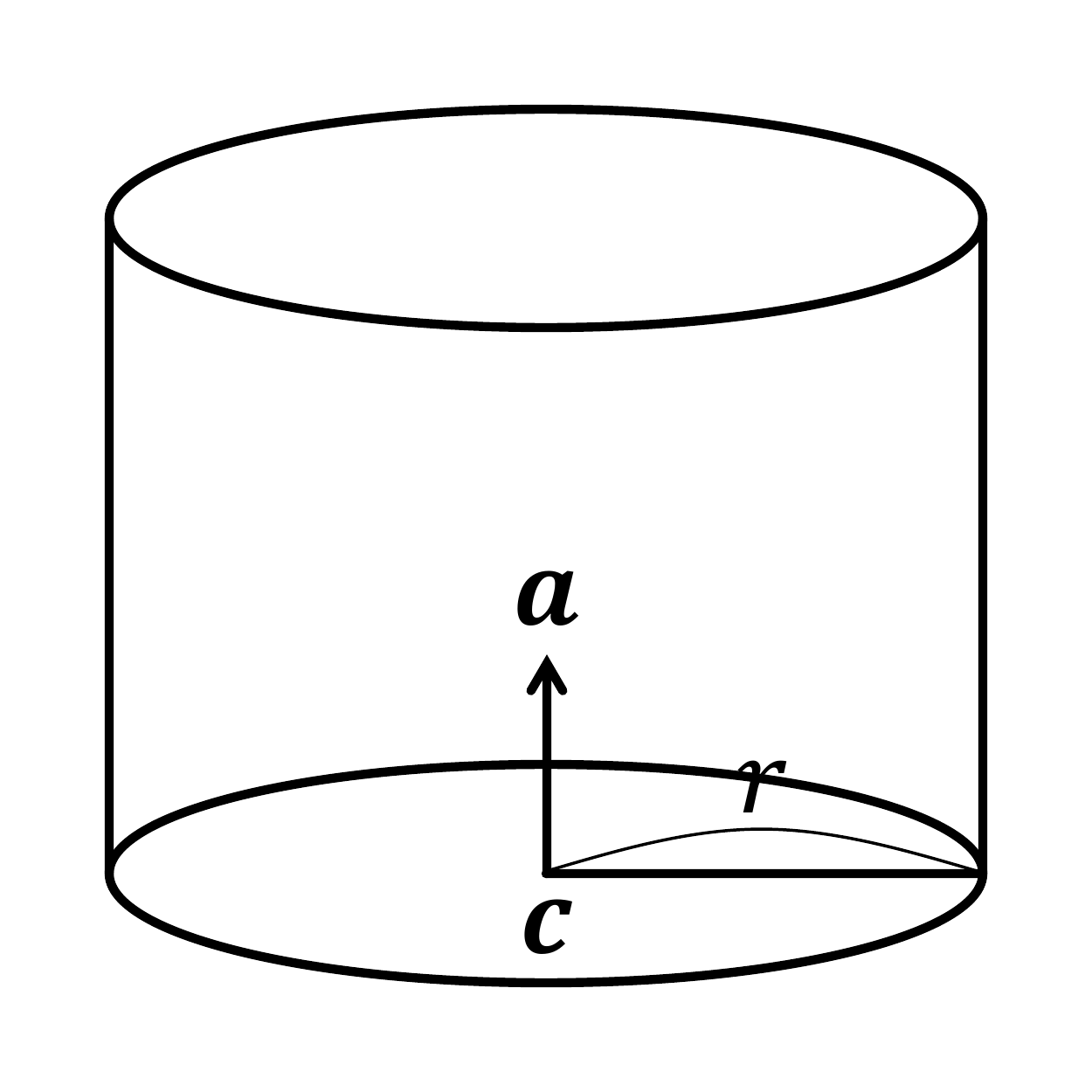}
\end{minipage}%
\begin{minipage}[t]{.25\columnwidth}
  \includegraphics[width=\linewidth,trim={0.8cm 0.8cm 0.8cm 0.8cm},clip]{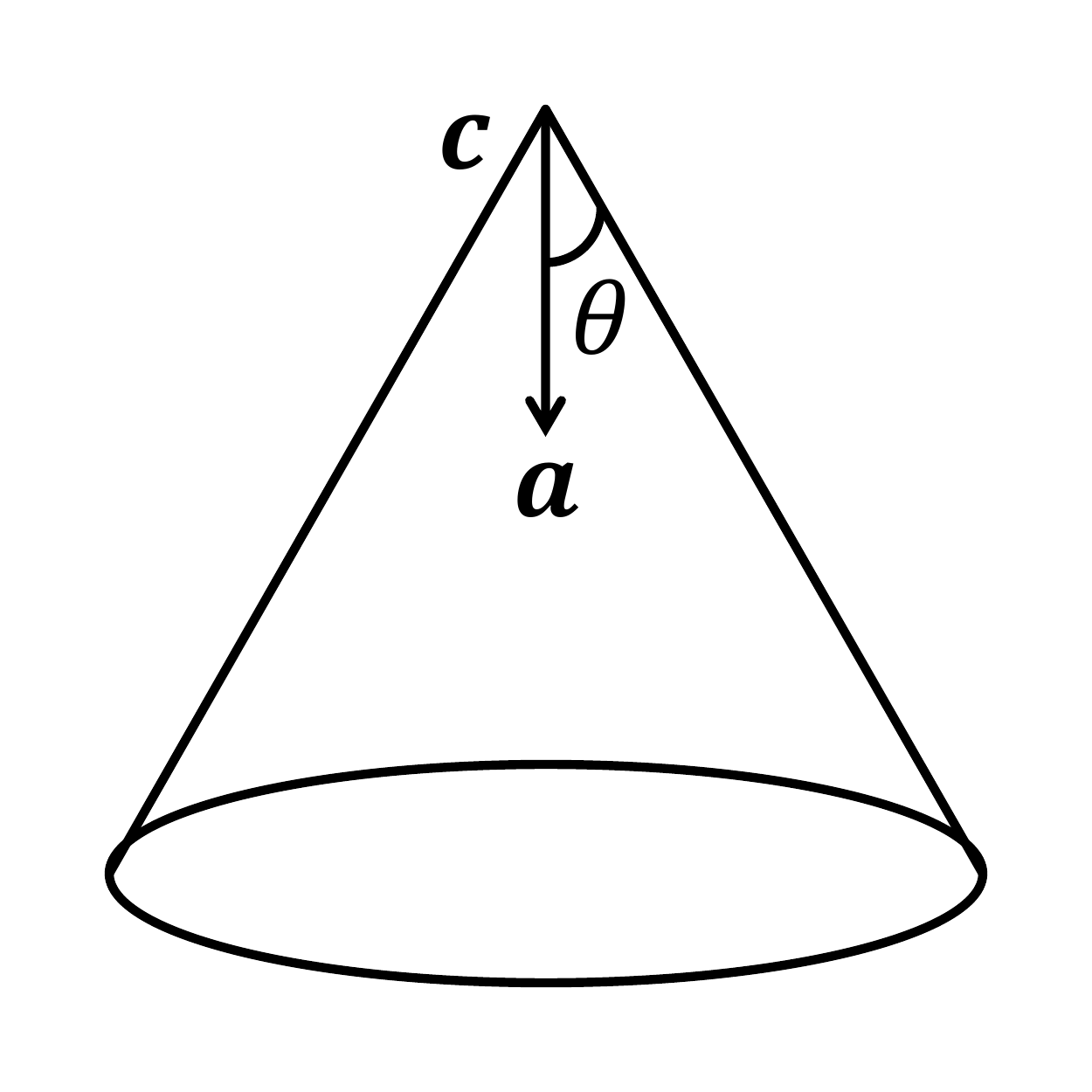}
\end{minipage}
\vspace{-\baselineskip}
\caption{Primitive types and parameters. The boundary information in each  $\mathbf{S}_k$, together with parameters $\mathbf{A}_k$, defines the (bounded) region of the primitive $k$. On the other hand, the point segment $\mathbf{W}_{:,k}$ provides an approximation to this bounded region.}
\label{fig:primitives}
\vspace{-\baselineskip}
\end{figure}

We propose Supervised Primitive Fitting Network (SPFN) that takes an input shape represented by a point cloud $\mathbf{P} \in \mathbb{R}^{N \times 3}$, where $N$ is number of points, and predicts a set of geometric primitives that best fit the input. 
The output of SPFN contains the type and parameters for every primitive, plus a list of input points assigned to it. 
Our network supports $L=4$ types of primitives: plane, sphere, cylinder, and cone (Figure~\ref{fig:primitives}), and we index these types by $0, 1, 2, 3$ accordingly.
Throughout the paper, we will use notations $\{\cdot\}_{i,:}$ and $\{\cdot\}_{:,k}$ to denote $i$-th row and $k$-th column of a matrix, respectively. 

During training, for each input shape with $K$ primitives, SPFN leverages the following ground truth information as supervision: point-to-primitive membership matrix $\mathbf{W} \in \{0, 1\}^{N\times K}$, unoriented point normals $\mathbf{N} \in \mathbb{R}^{N \times 3}$, and bounded primitive surfaces $\{\mathbf{S}_{k}\}_{k=1,\ldots, K}$. 
For the membership matrix, $\mathbf{W}_{i,k}$ indicates if point $i$ belongs to primitive $k$ so that $\sum_{k=1}^K \mathbf{W}_{i,k} \le 1$.
Notice that $\mathbf{W}_{:,k}$, the $k$-th column of $\mathbf{W}$, indicates the \emph{point segment} assigned to primitive $k$. 
We allow $K$ to \emph{vary} for each shape, and $\mathbf{W}$ can have zero rows indicating \emph{unassigned} points (points not belonging to any of the $K$ primitives; e.g. it belongs to a primitive of unknown type).
Each $\mathbf{S}_k$ contains information about the type, parameters, and boundary of the $k$-th primitive surface, and we denote its type by $\mathbf{t}_k \in \{0,1,\ldots, L-1\}$ and its type-specific parameters by $\mathbf{A}_k$.
We include the boundary of $\mathbf{S}_k$ in the supervision besides $\mathbf{P}$ because $\mathbf{P}$ can be noisy, and we do not discriminate against small surfaces in evaluating per-primitive losses (see Equation \ref{eq:residual_loss}).
For convenience, we define per-point type matrix $\mathbf{T} \in \{0, 1\}^{N \times L}$ by $\mathbf{T}_{i,l} = \sum_{k=1}^K \mathbb{1} (\mathbf{W}_{i,k}=1)\mathbb{1}(\mathbf{t}_k = l)$, where $\mathbb{1}(\cdot)$ is the indicator function.

The pipeline of SPFN at training time is illustrated in Figure~\ref{fig:pipeline}. We use PointNet++~\cite{Qi:2017b} segmentation architecture to consume the input point cloud $\mathbf{P}$. A slight modification is that we add three separate fully-connected layers to the end of the PointNet++ pipeline in order to predict the following per-point properties: point-to-primitive membership matrix $\mathbf{\hat{W}} \in [0, 1]^{N\times K}$\footnote{For notational clarity, for now we assume the number of predicted primitives equals $K$, the number of ground truth primitives. See Section \ref{sec:implementation_details} for how to predict $\mathbf{\hat W}$ without prior knowledge of $K$.}, unoriented point normals $\mathbf{\hat{N}} \in \mathbb{R}^{N \times 3}$, and per-point primitive types $\mathbf{\hat{T}} \in [0, 1]^{N\times L}$. We use softmax activation to obtain membership probabilities in the rows of $\mathbf{\hat{W}}$ and $\mathbf{\hat{T}}$, and we normalize the rows of the $\mathbf{\hat{N}}$ to constrain normals to have $l^2$-norm $1$. 
We then feed these per-point quantities to our \emph{differentiable model estimator} (Section \ref{sec:model_estimation}) that computes primitive parameters $\{\mathbf{\hat{A}}_k\}$ based on the per-point information. Since this last step is differentiable, we are able to backpropagate any kind of per-primitive loss through the PointNet++, and thus the training can be done end-to-end.


Notice that we do not assume a consistent ordering of ground truth primitives, so we do not assume any ordering of the columns of our predicted $\mathbf{\hat{W}}$.
In Section~\ref{sec:primitives_reordering}, we describe the \emph{primitive reordering} step used to handle such mismatch of orderings. In Section~\ref{sec:model_estimation}, we present our differential model estimator for predicting primitive parameters $\{\mathbf{\hat{A}}_{k}\}$. In Section~\ref{sec:loss_function}, we define each term in our loss function. Lastly, in Section~\ref{sec:implementation_details}, we describe implementation details.

\subsection{Primitives Reordering}
\label{sec:primitives_reordering}
\vspace{-3pt}
Inspired by Yi \etal~\cite{Yi:2018}, we compute \emph{Relaxed Intersection over Union} (RIoU)~\cite{Krahenbuhl:2013} for all pairs of columns from the membership matrices $\mathbf{W}$ and $\mathbf{\hat{W}}$. The RIoU for two indicator vectors $\mathbf{w}$ and $\mathbf{\hat{w}}$ is defined as follows:
\begin{align}
\text{RIoU}(\mathbf{w}, \mathbf{\hat{w}}) = \frac{\mathbf{w}^\mathbf{T}\mathbf{\hat{w}}}{\|\mathbf{w}\|_1 + \|\mathbf{\hat{w}}\|_1 -  \mathbf{w}^\mathbf{T}\mathbf{\hat{w}}}.
\end{align}
The best one-to-one correspondence (determined by RIoU) between columns of the two matrices is then given by Hungarian matching~\cite{Kuhn:1955}. 
We reorder the ground truth primitives according to this correspondence, so that ground truth primitive $k$ is matched with the predicted primitive $k$. 
Since the set of inputs where a small perturbation will lead to a change of the matching result has measure zero, the overall pipeline remains differentiable almost everywhere. Hence we use an external Hungarian matching solver to obtain optimal matching indices, and then inject these back into our network to allow further loss computation and gradient propagation.

\subsection{Primitive Model Estimation}
\label{sec:model_estimation}
\vspace{-3pt}
In the model estimation module, primitive parameters $\{\mathbf{A}_k\}$ are obtained from the predicted per-point properties in a differentiable manner. 
As the parameter estimation for each primitive is independent, in this section we will assume $k$ is a fixed index of a primitive.
The input to the model estimation module consists of $\mathbf{P}$, the input point cloud, $\mathbf{\hat{N}}$, the predicted unoriented point normals, and $\mathbf{\hat{W}}_{:,k}$, the $k$-th column of the predicted membership matrix $\mathbf{\hat{W}}$. For simplicity, we write $\mathbf{w} = \mathbf{\hat{W}}_{:,k} \in [0, 1]^N$ and $\mathbf{\hat{A}} = \mathbf{\hat{A}}_k$. 
For $\mathbf{p} \in \mathbb{R}^3$, let $D_{l}(\mathbf{p}, \mathbf{A})$ denote the distance from $\mathbf{p}$ to the primitive of type $l$ and parameters $\mathbf{A}$.
The differentiable module for computing $\mathbf{\hat{A}}$, given the primitive type, is illustrated below.

\vspace{-\baselineskip}
\paragraph{Plane.}
A plane is represented by $\mathbf{A} = (\mathbf{a}, d)$ where $\mathbf{a}$ is the normal of the plane, with $\| \mathbf{a} \| = 1$, and the points on the plane are $\{\mathbf{p} \in \mathbb{R}^3 : \mathbf{a}^{\mathbf{T}} \mathbf{p} = d\}$. Then
\begin{align}
  D^2_{\text{plane}}(\mathbf{p}, \mathbf{A}) = (\mathbf{a}^{\mathbf{T}} \mathbf{p} - d)^2.
\label{eq:plane_residual}
\end{align}
We can then define $\mathbf{\hat{A}}$ as the minimizer to the \emph{weighted} sum of squared distances as a function of $\mathbf{A}$:
\begin{align}
  \mathcal{E}_{\text{plane}}(\mathbf{A}; \mathbf{P}, \mathbf{w})
  &= \sum_{i=1}^N \mathbf{w}_i (\mathbf{a}^\mathbf{T} \mathbf{P}_{i,:} - d)^2.
\label{eq:plane_weighted}
\end{align}
By solving $\frac{\partial \mathcal{E}_{\text{plane}}}{\partial d} = 0$, we obtain $d = \frac{\sum_{i=1}^N \mathbf{w}_i \mathbf{a}^\mathbf{T} \mathbf{P}_{i,:}}{\sum_{i=1}^N \mathbf{w}_i}$.
Plugging this into Equation~$\ref{eq:plane_weighted}$ gives:
\begin{equation}
  \mathcal{E}_{\text{plane}}(\mathbf{a}; \mathbf{P}, \mathbf{w}) = \norm{\left(\text{diag}\left(\mathbf{w}\right)\right)^\frac{1}{2} \mathbf{X} \mathbf{a}}^2,
\label{eq:plane_weighted_axis}
\end{equation}
where $\mathbf{X}_{i,:} = \mathbf{P}_{i,:} - \textstyle\frac{\sum_{i=1}^N \mathbf{w}_i \mathbf{P}_{i,:}}{\sum_{i=1}^N \mathbf{w}_i}$.
Hence minimizing $\mathcal{E}_{\text{plane}}(\mathbf{A}; \mathbf{P}, \mathbf{w})$ over $\mathbf{a}$ becomes a homogeneous least square problem subject to $\| \mathbf{a} \| = 1$, and its solution is given as the right singular vector $\mathbf{v}$ corresponding to the smallest singular value of matrix $\left(\text{diag}\left(\mathbf{w}\right)\right)^\frac{1}{2} \mathbf{X}$. As shown by Ionescu~\etal~\cite{Ionescu:2015a,Ionescu:2015b}, the gradient with respect to $\mathbf{v}$ can be backpropagated through the SVD computation.

\vspace{-\baselineskip}
\paragraph{Sphere.}
A sphere is parameterized by $\mathbf{A} = (\mathbf{c}, r)$, where $\mathbf{c} \in \mathbb{R}^3$ is the center and $r \in \mathbb{R}$ is the radius. Hence
\begin{align}
  D^2_{\text{sphere}}(\mathbf{p}, \mathbf{A}) = (\norm{\mathbf{p} - \mathbf{c}} - r)^2
.\end{align}
In the sphere case (also in the cases of cylinder and cone), the squared distance is not quadratic. Hence minimizing the weighted sum of squared distances over parameters as done in the plane is only available via nonlinear iterative solvers~\cite{Lukacs:1998}. Instead, we consider minimizing over the weighted sum of a different notion of distance:
\begin{align}
  \mathcal{E}_{\text{sphere}}(\mathbf{A}; \mathbf{P}, \mathbf{w}) = \sum_{i=1}^N \mathbf{w}_i (\norm{\mathbf{P}_{i,:} - \mathbf{c}}^2 -r^2)^2.
  \label{eq:sphere_weighted}
\end{align}
Solving $\frac{\partial \mathcal{E}_{\text{sphere}}}{\partial r^2} = 0$ gives $r^2 = \frac{1}{\sum_{i=1}^N \mathbf{w}_i} \sum_{j=1}^N\mathbf{w}_j \|\mathbf{P}_j - \mathbf{c}\|^2$. 
Putting this back in Equation \ref{eq:sphere_weighted}, we end up with a quadratic expression in $\mathbf{c}$ as a least square:
\begin{equation}
  \mathcal{E}_{\text{sphere}}(\mathbf{c}; \mathbf{P}, \mathbf{w}, \mathbf{a}) = \norm{\left(\text{diag}\left(\mathbf{w}\right)\right)^\frac{1}{2} (\mathbf{X}\mathbf{c} -\mathbf{y})}^2,
\label{eq:sphere_weighted_center}
\end{equation}
where $\mathbf{X}_{i,:} = 2\left(-\mathbf{P}_{i,:} + \textstyle\frac{\sum_{j=1}^N \mathbf{w}_j \mathbf{P}_{j,:}}{\sum_{j=1}^N \mathbf{w}_j}\right)$ and
$\mathbf{y}_{i} = \mathbf{P}_{i,:}^{\mathbf{T}}\mathbf{P}_{i,:} - \textstyle\frac{\sum_{j=1}^N \mathbf{w}_j \mathbf{P}_{j,:}^{\mathbf{T}}\mathbf{P}_{j,:}}{\sum_{j=1}^N \mathbf{w}_j}$.
This least square can be solved via Cholesky factorization in a differentiable way~\cite{Murray:2016}.

\vspace{-\baselineskip}
\paragraph{Cylinder.}
A cylinder is parameterized by $\mathbf{A} = (\mathbf{a}, \mathbf{c}, r)$ where $\mathbf{a} \in \mathbb{R}^3$ is a unit vector of the axis, $\mathbf{c} \in \mathbb{R}^3$ is the center, and $r \in \mathbb{R}$ is the radius. We have
\begin{equation}
  D^2_{\text{cylinder}}(\mathbf{p}, \mathbf{A}) = \left( \sqrt{\mathbf{v}^{\mathbf{T}}\mathbf{v} - (\mathbf{a}^{\mathbf{T}}\mathbf{v})^2} - r \right)^2,
\end{equation}
where $\mathbf{v} = \mathbf{p} - \mathbf{c}$.
As in the sphere case, directly minimizing over squared true distance is challenging. Instead, inspired by Nurunnabi et. al.~\cite{Nurunnabi:2017}, we first estimate the axis $\mathbf{a}$ and then solve a circle fitting to obtain the rest of the parameters.
Observe that the normals of points on the cylinder must be perpendicular to $\mathbf{a}$, so we choose $\mathbf{a}$ to minimize:
\begin{align}
  \mathcal{E}_{\text{cylinder}}(\mathbf{a}; \mathbf{\hat{N}}, \mathbf{w}) = \norm{ \left(\text{diag}\left(\mathbf{w}\right)\right)^\frac{1}{2} \mathbf{\hat{N}} \mathbf{a} }^2,
  \label{eq:cylinder_axis}
\end{align}
which is a homogeneous least square problem same as Equation~\ref{eq:plane_weighted_axis}, and can be solved in the same way.

Once obtaining the axis $\mathbf{a}$, we consider a plane $\mathscr{P}$ with normal $\mathbf{a}$ that passes through the origin, and notice the projection of the cylinder onto $\mathscr{P}$ should form a circle. 
Thus we can choose $\mathbf{c}$ and $r$ to be the circle that best fits the projected points $\{\text{Proj}_\mathbf{a}(\mathbf{P}_{i,:})\}_{i=1}^N$,
where $\text{Proj}_\mathbf{a}(\cdot)$ denotes the projection onto $\mathscr{P}$. This is exactly the same formulation as in the sphere case (Equation~\ref{eq:sphere_weighted}), and can thus be solved similarly. 

\vspace{-\baselineskip}
\paragraph{Cone.}
A cone is parameterized by $\mathbf{A} = (\mathbf{a}, \mathbf{c}, \theta)$ where $\mathbf{c} \in \mathbb{R}^3$ is the apex, $\mathbf{a} \in \mathbb{R}^3$ is a unit vector of the axis from the apex into the cone, and $\theta \in (0, \frac{\pi}{2})$ is the half angle. Then
\begin{equation}
  D^2_{\text{cone}}(\mathbf{p}, \mathbf{A})^2 = \left( \|\mathbf{v}\| \sin\left( \min\left( \left| \alpha - \theta \right|, \frac{\pi}{2}  \right) \right) \right)^2,
\end{equation}
where $\mathbf{v} = \mathbf{p} - \mathbf{c}, 
  \alpha = \arccos\left( \frac{\mathbf{a}^{\mathbf{T}}\mathbf{v}}{\|\mathbf{v}\|}\right)$.
Similarly with the cylinder case, we use a multi-stage algorithm: first we estimate $\mathbf{a}$ and $\mathbf{c}$ separately, and then we estimate the half-angle $\theta$.

We utilize the fact that the apex $\mathbf{c}$ must be the intersection point of all tangent planes on the cone surface. Using the predicted point normals $\mathbf{\hat{N}}$, the multi-plane intersection problem is formulated as a least square similar with Equation~\ref{eq:sphere_weighted_center} by minimizing
\begin{align}
 \mathcal{E}_{\text{cone}}(\mathbf{c}; \mathbf{\hat{N}}) = \norm{ \left(\text{diag}\left(\mathbf{w}\right)\right)^\frac{1}{2} \left( \mathbf{\hat{N}} \mathbf{c} - \mathbf{y} \right) }^2,
\label{eq:cone_apex}
\end{align}
where $\mathbf{y}_{i} = \mathbf{\hat{N}}_{i,:}^{\mathbf{T}}\mathbf{P}_{i,:}$.
To get the axis direction $\mathbf{a}$, observe that $\mathbf{a}$ should be the normal of the plane passing through all $\mathbf{N}_i$ if point $i$ belongs to the cone.
This is just a plane fitting problem, and we can compute $\mathbf{a}$ as the unit normal that minimizes Equation \ref{eq:plane_weighted}, where we replace $\mathbf{P}_{i,:}$ by $\mathbf{\hat{N}}_{i,:}$. 
We flip the sign of $\mathbf{a}$ if it is not going from $\mathbf{c}$ into the cone.
Finally, using the apex $\mathbf{c}$ and the axis $\mathbf{a}$, the half-angle $\theta$ is simply computed as a weighted average:
\begin{align}
  \theta = \textstyle\frac{1}{\sum_{i=1}^N \mathbf{w}_i} \sum_{i=1}^N \mathbf{w}_i \arccos \left|\mathbf{a}^\mathbf{T} \textstyle\frac{\mathbf{P}_{i,:}-\mathbf{c}}{\norm{\mathbf{P}_{i,:} - \mathbf{c}}}\right|.
\end{align}

\subsection{Loss Function}
\label{sec:loss_function}
\vspace{-3pt}

We define our loss function $\mathcal{L}$ as the sum of the following five terms without weights:
\begin{align}
  \mathcal{L} = \mathcal{L}_{\text{seg}} + \mathcal{L}_{\text{norm}} + \mathcal{L}_{\text{type}} + \mathcal{L}_{\text{res}} + \mathcal{L}_{\text{axis}}.
\end{align}
Each loss term is described below for a single input shape.

\vspace{-\baselineskip}
\paragraph{Segmentation Loss.}
The primitive parameters can be more accurately estimated when the segmentation of the input point cloud is close to the ground truth. Thus, we minimize $(1 - \text{RIoU})$ for each pair of a ground truth primitive and its correspondence in the prediction:
\begin{align}
  \mathcal{L}_{\text{seg}} = \frac{1}{K} \sum_{k=1}^{K} \left( 1 - \text{RIoU}(\mathbf{W}_{:,k}, \mathbf{\hat{W}}_{:, k}) \right).
\end{align}

\vspace{-\baselineskip}
\paragraph{Point Normal Angle Loss.}
For predicting the point normals $\mathbf{\hat{N}}$ accurately, we minimize the absolute cosine angle between ground truth and predicted normals:
\begin{align}
  \mathcal{L}_{\text{norm}} = \frac{1}{N} \sum_{i=1}^{N} \left( 1 - |\mathbf{N}_{i, :}^\mathbf{T}\mathbf{\hat{N}}_{i, :}| \right).
\end{align}
The absolute value is taken since our predicted normals are unoriented.

\vspace{-\baselineskip}
\paragraph{Per-point Primitive Type Loss.}
We minimize cross entropy $H$ for the per-point primitive types $\mathbf{\hat{T}}$ (unassigned points are ignored):
\begin{align}
  \mathcal{L}_{\text{type}} = \frac{1}{N} \sum_{i=1}^{N} \mathbb{1}(\mathbf{W}_{i,:} \neq \mathbf{0}) H(\mathbf{T}_{i,:}, \mathbf{\hat{T}}_{i,:}),
\end{align}
where $\mathbb{1}(\cdot)$ is the indicator function.

\vspace{-\baselineskip}
\paragraph{Fitting Residual Loss.}
Most importantly, we minimize the expected squared distance between $\mathbf{S}_k$ and the predicted primitive $k$ parameterized by $\mathbf{\hat{A}}_{k}$ across all $k = 1,\ldots, K$:
\begin{align}
  \mathcal{L}_{\text{res}} &= \frac{1}{K} \sum_{k=1}^{K} \mathbb{E}_{\mathbf{p} \sim U(\mathbf{S}_k)} D^2_{\mathbf{t}_k}(\mathbf{p}, \mathbf{\hat{A}}_{k}),
\label{eq:residual_loss}
\end{align}
where $\mathbf{p}\sim U(\mathbf{S})$ means $\mathbf{p}$ is sampled uniformly on the bounded surface $\mathbf{S}$ when taking the expectation, and $D^2_{l}(\mathbf{p}, \mathbf{\hat{A}})$ is the squared distance from $\mathbf{p}$ to a primitive of type $l$ with parameter $\mathbf{\hat A}$, as defined in Section~\ref{sec:model_estimation}. 
Note that every $\mathbf{S}_k$ is weighted equally in Equation~\ref{eq:residual_loss} regardless of its \emph{scale}, the surface area relative to the entire shape. This allows us to detect small primitives that can be missed by other unsupervised methods.

Note that in Equation \ref{eq:residual_loss}, we use the ground truth type $\mathbf{t}_k$ instead of inferring the predicted type based on $\mathbf{\hat{T}}$ and then properly weighted by $\mathbf{\hat{W}}$. 
We do this because coupling multiple predictions can make loss functions more complicated, resulting in unstable training.
At test time, however, the type of primitive $k$ is predicted as
\begin{align}
\mathbf{\hat t}_k = \argmax_l \sum_{i=1}^N \mathbf{\hat{T}}_{i,l} \mathbf{\hat{W}}_{i,k}.
\label{eq:pred_per_primitive_type}
\end{align}

\vspace{-\baselineskip}
\paragraph{Axis Angle Loss.}
Estimating plane normal and cylinder/cone axis using SVD can become numerically unstable when the predicted $\mathbf{\hat{W}}$ leads to degenerate cases, such as when the number of points with a nonzero weight is too small, or when the points with substantial weights form a narrow plane close to a line during plane normal estimation (Equation \ref{eq:plane_weighted_axis}). Thus, we regularize the axis parameters with a cosine angle loss:
\begin{align}
  \mathcal{L}_{\text{axis}} = \frac{1}{K} \sum_{k=1}^K \left(1 - \Theta_{\mathbf{t}_k}(\mathbf{A}_k, \mathbf{\hat{A}}_k) \right),
\end{align}
where $\Theta_{t}(\mathbf{A}, \mathbf{\hat{A}})$ denotes $|\mathbf{a}^\mathbf{T}\mathbf{\hat{a}}|$ for plane (normal), cylinder (axis), and cone (axis), and $1$ for sphere (so the loss becomes zero).

\subsection{Implementation Details}
\label{sec:implementation_details}
\vspace{-3pt}
In our implementation, we assume a fixed number $N$ \rev{of input points} for all shapes.
While the number of ground truth primitives varies across the input shapes, we choose an integer $K_\text{max}$ in prediction to fix the size the output membership matrix $\mathbf{\hat{W}} \in \mathbb{R}^{N\times K_\text{max}}$ so that $K_\text{max}$ is no less than the maximum primitive numbers in input shapes. After the Hungarian matching in Section~\ref{sec:primitives_reordering}, unmatched columns in $\mathbf{\hat{W}}$ are ignored in the loss computation.
At test time, we discard a predicted primitive $k$ if 
$
    \frac{\sum_{i=1}^N \mathbf{\hat{W}}_{i,k}}{N} > \epsilon_\text{discard}
    $,
\rev{where $\epsilon_\text{discard}=0.005N$ for all experiments. This is just a rather arbitrary small threshold to weed out unused segments.}

When evaluating the expectation $\mathbb{E}_{\mathbf{p} \sim U(\mathbf{S}_k)}(\cdot)$ in Equation \ref{eq:residual_loss}, on-the-fly point sampling takes very long time in training.
Hence the expectation is approximated as the average for $M$ points on $\mathbf{S}_k$ that are sampled uniformly when preprocessing the data.


\section{Experiments}
\label{sec:experiments}
\vspace{-3pt}


\begin{table*}[t!]
\centering
\newcolumntype{Y}{>{\centering\arraybackslash}X}
\footnotesize{
{
\setlength{\tabcolsep}{0.2em}
\renewcommand{\arraystretch}{0.9}
\begin{tabularx}{\textwidth}{l|m{3.1cm}|Y|Y|Y|Y|>{\centering}m{1.8cm}|Y|Y|Y|Y}
  \toprule
    \multirow{2}{*}{\scriptsize{Ind}} &
    \multirow{2}{*}{Method} &
    \multirow{2}{*}{{\makecell{Seg.\\(Mean IoU)}}} &
    \multirow{2}{*}{{\makecell{Primitive\\Type (\%)}}} &
    \multirow{2}{*}{{\makecell{Point\\Normal ($^{\circ}$)}}} &
    \multirow{2}{*}{{\makecell{Primitive\\Axis ($^{\circ}$)}}} &
    \multirow{2}{*}{{\makecell{$\{\mathbf{S}_k\}$ Residual\\Mean $\pm$ Std.}}} &
    \multicolumn{2}{c|}{$\{\mathbf{S}_k\}$ Coverage} &
    \multicolumn{2}{c}{$\mathbf{P}$  Coverage} \\
  \cline{8-11}
     & & & & & & &
     $\epsilon = 0.01$ & $\epsilon = 0.02$ & $\epsilon = 0.01$ & $\epsilon = 0.02$ \\
  \midrule
  1 & \scriptsize{Eff. RANSAC~\cite{Schnabel:2007}}+J &
  43.68 & 52.92 & 11.42 & 7.54 & 0.072 $\pm$ 0.361 & 43.42 & 63.16 & 65.74 & 88.63  \\
  2 & \scriptsize{Eff. RANSAC~\cite{Schnabel:2007}}*+J* &
  56.07 & 43.90 & 6.92 & 2.42 & 0.067 $\pm$ 0.352 & 56.95 &  72.74  & 68.58  &  92.41  \\
  3 & \scriptsize{Eff. RANSAC~\cite{Schnabel:2007}}+J* &
  45.90 & 46.99 & \textbf{6.87} & 5.85 & 0.080 $\pm$ 0.390 & 51.59 &  67.12 & 72.11 & 92.58 \\
  4 & \scriptsize{Eff. RANSAC~\cite{Schnabel:2007}}+J*+$\mathbf{\hat{W}}$ &
  69.91 & 60.56 & \textbf{6.87} & 2.90 & 0.029 $\pm$ 0.234 & 74.32 &  83.27  & 78.79  &  94.58  \\
  5 & \scriptsize{Eff. RANSAC~\cite{Schnabel:2007}}+J*+$\mathbf{\hat{W}}$+$\mathbf{\hat{t}}$ &
  60.68 & 92.76 & \textbf{6.87} & 6.21 & 0.036 $\pm$ 0.251 & 65.31 &  73.69  & 77.01  &  92.57  \\
  6 & \scriptsize{Eff. RANSAC~\cite{Schnabel:2007}}+$\mathbf{\hat{N}}$+$\mathbf{\hat{W}}$+$\mathbf{\hat{t}}$ &
  60.56 & 93.13 & 8.15 & 7.02 & 0.054 $\pm$ 0.307 & 61.94 &  70.38 & 74.80 & 90.83 \\
  \midrule
  7 & DPPN (Sec. 4.4) & 44.05 & 51.33 & - & 3.68 & 0.021 $\pm$ 0.158 & 46.99 & 71.02 & 59.74 & 84.37 \\
  \midrule
  8 & SPFN-$\mathcal{L}_{\text{seg}}$ &
  41.61 & 92.40 & 8.25 & 1.70 & 0.029 $\pm$ 0.178 & 50.04 & 62.74 & 62.23 & 77.74 \\
  9 & SPFN-$\mathcal{L}_{\text{norm}}$+J* &
  71.18 & 95.44 & \textbf{6.87} & 4.20 & 0.022 $\pm$ 0.188 & 76.47 & 81.49 & 83.21 & 91.73 \\
  10 & SPFN-$\mathcal{L}_{\text{res}}$ &
  72.70 & 96.66 & 8.74 & 1.87 & 0.017 $\pm$ 0.162 & 79.81 & 85.57 & 81.32 & 91.52 \\
  11 & SPFN-$\mathcal{L}_{\text{axis}}$ &
  \textbf{77.31} & 96.47 & 8.28 & 6.27 & 0.019 $\pm$ 0.188 & 80.80 & 86.11 & 86.46 & 94.43 \\
  12 & SPFN ($\mathbf{\hat{t}} \rightarrow$ Est.) &
  75.71 & 95.95 & 8.54 & 1.71 & 0.013 $\pm$ 0.140 & 85.25 & 90.13 & 86.67 & 94.91 \\
  13 & SPFN &
  77.14 & \textbf{96.93} & 8.66 & \textbf{1.51} & \textbf{0.011 $\pm$ 0.131} & \textbf{86.63} & \textbf{91.64} & \textbf{88.31} & \textbf{96.30}  \\
  \bottomrule
\end{tabularx}
}
}
\vspace{-0.5\baselineskip}
\caption{Results of all experiments. +J indicates using point normals computed by jet fitting~\cite{Cazals:2003} from the input point clouds. The asterisk * indicates using high resolution ($64k$) point clouds. See Section~\ref{sec:evaluation_metrics} for the details of evaluation metrics, and \cref{sec:efficient_ransac,sec:dppn,sec:ablation_study} for the description of each experiment. Lower is better in 3-5\textsuperscript{th} metrics, and higher is better in the rest.}
\label{tbl:results}
\end{table*}


\begin{figure*}[t!]
\centering
\newcolumntype{Y}{>{\centering\arraybackslash}X}
\footnotesize{
{
\setlength{\tabcolsep}{0.0em}
\renewcommand{\arraystretch}{0.9}
\begin{tabularx}{\textwidth}{m{0.05\textwidth}m{0.95\textwidth}}
  \scriptsize{\makecell{Ground\\Truth}} &
  \includegraphics[width=0.95\textwidth]{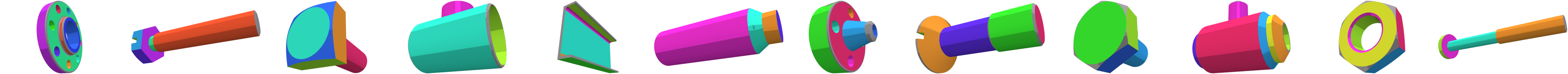} \\
  \midrule
  \scriptsize{\makecell{Eff.RAN.\\+J}} &
  \includegraphics[width=0.95\textwidth]{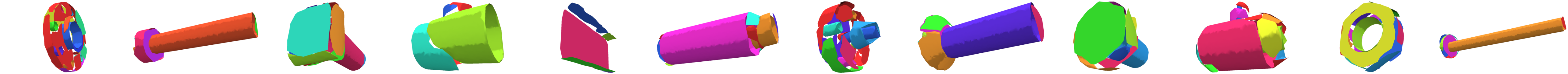} \\
  \scriptsize{\makecell{Eff.RAN.*\\+J*}} &
  \includegraphics[width=0.95\textwidth]{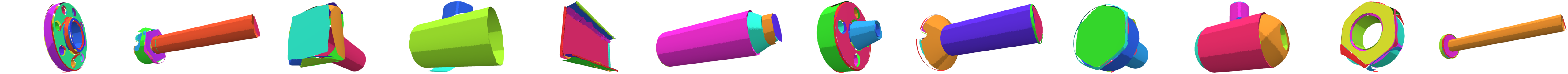} \\
  \scriptsize{\makecell{Eff.RAN+\\$\mathbf{\hat{N}}$+$\mathbf{\hat{W}}$+$\mathbf{\hat{t}}$}} &
  \includegraphics[width=0.95\textwidth]{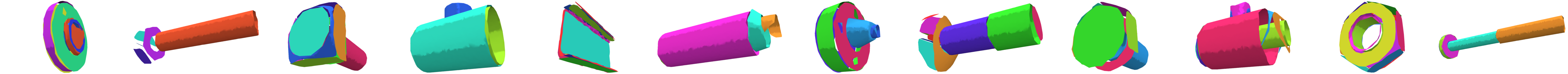} \\
  \scriptsize{\makecell{DPPN}} & \includegraphics[width=0.95\textwidth]{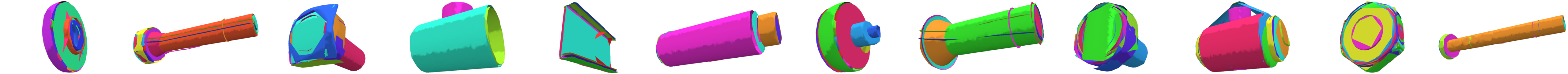} \\
  \scriptsize{\makecell{SPFN\\-$\mathcal{L}_{\text{seg}}$}} & \includegraphics[width=0.95\textwidth]{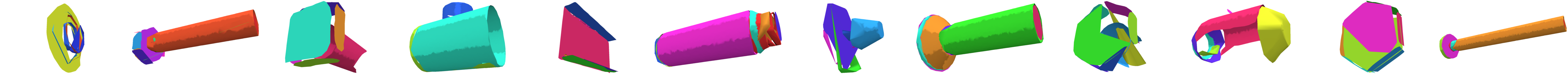} \\
  \scriptsize{\makecell{SPFN\\-$\mathcal{L}_{\text{norm}}$+J*}} & \includegraphics[width=0.95\textwidth]{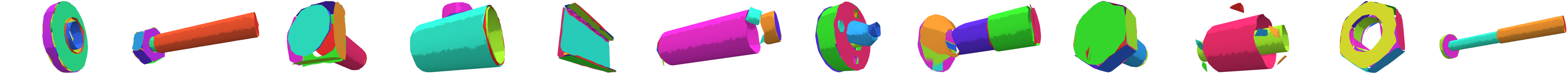} \\
  \scriptsize{\makecell{SPFN\\-$\mathcal{L}_{\text{res}}$}} & \includegraphics[width=0.95\textwidth]{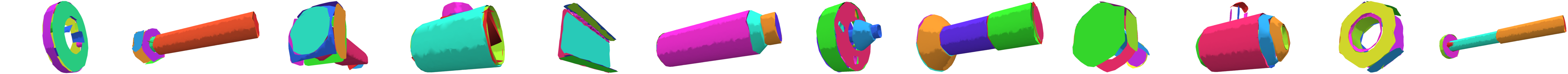} \\
  \scriptsize{\makecell{SPFN\\-$\mathcal{L}_{\text{axis}}$}} & \includegraphics[width=0.95\textwidth]{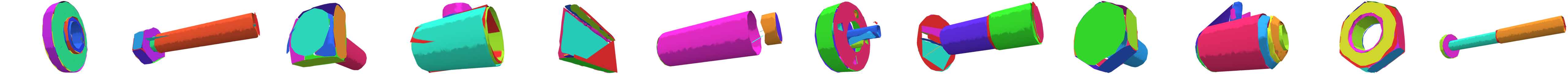} \\
  \scriptsize{\makecell{SPFN}} & \includegraphics[width=0.95\textwidth]{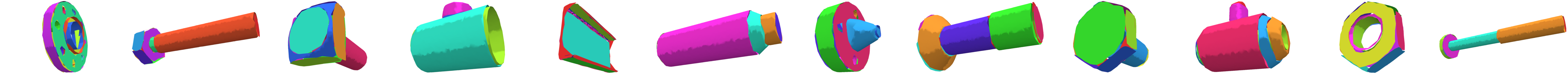}
\end{tabularx}
}
}
\vspace{-0.5\baselineskip}
\caption{Primitive fitting results with different methods. The results are rendered with meshes generated by projecting point segments to output primitives and then triangulating them. Refer to \cref{sec:efficient_ransac,sec:dppn,sec:ablation_study} for the details of each method.}
\label{fig:results}
\vspace{-1\baselineskip}
\end{figure*}




\subsection{ANSI Mechanical Component Dataset}
\label{sec:dataset}
\vspace{-3pt}
For training and evaluating the proposed network, we use CAD models from American National Standards Institute (ANSI)~\cite{ANSI} mechanical components provided by TraceParts~\cite{TraceParts}. Since there is no existing \emph{scanned} 3D dataset for this type of objects, we train and test our network by generating noisy samples on these models. 
From 504 categories, we randomly select up to 100 models in each category for balance and diversity, and split training/test sets \emph{by categories} so that training and test models are from disjoint categories, resulting in 13,831/3,366 models in training/test sets. 
We remark that the four types of primitives we consider (plane, sphere, cylinder, cone) cover $94.0\%$ percentage of area per-model on average in our dataset. 
When generating the point samples from models, we still include surfaces that are not one of the four types.
The maximum number of primitives per shape does not exceed $20$ in all our models. We set $K_\text{max}=24$ where we add $4$ extra columns in $\mathbf{\hat{W}}$ to allow the neural net to assign a small number of points to the extra columns, effectively marking those points unassigned because of  the threshold $\epsilon_\text{discard}$.

From the CAD models, we extract primitives information including their boundaries. 
We then merge adjacent pieces of primitive surfaces sharing exactly the same parameters; this happens because of the difficulty of representing boundaries in CAD models, so for instance a complete cylinder will be split into a disjoint union of two mirrored half cylinders.
We discard tiny pieces of primitives (less than 2\% of the entire area). 
Each shape is normalized so that its center of mass is at the origin, and the axis-aligned bounding box for the shape is included in $[-1, 1]$ range along every axis.
In experiments, we first uniformly sample $8192$ points over the entire surface of each shape as the input point cloud ($N = 8192$). This is done by first sampling on the discretized mesh of the shape and then projecting all points onto its geometric surface. Then we randomly apply noise to the point cloud along the surface normal direction in $[-0.01, 0.01]$ range. To evaluate the fitting residual loss $\mathcal{L}_{\text{res}}$, we also uniformly sample 512 points per primitive surface for approximating $\mathbf{S}_k$ ($M=512$).

\subsection{Evaluation Metrics}
\label{sec:evaluation_metrics}
\vspace{-3pt}
We design our evaluation metrics as below. Each quantity is described for a single shape, and the numbers are reported as the average of these quantities across all test shapes. For per-primitive metrics, we first perform primitive reordering as in Section \ref{sec:primitives_reordering} so the indices for predicted and ground truth primitives are matched.

\begin{itemize}
  \setlength\itemsep{-0.1em}

  \item \textbf{Segmentation Mean IoU}:\\
    $\frac{1}{K} \sum_{k=1}^{K} \text{IoU}(\mathbf{W}_{:,k}, \mathcal{I}(\mathbf{\hat{W}}_{:, k}))$, where $\mathcal{I}(\cdot)$ is the one-hot conversion.

  \item \textbf{Mean primitive type accuracy}:\\
    $\frac{1}{K} \sum_{k=1}^{K} \mathbb{1}({\mathbf{t}_{k} = \mathbf{\hat{t}}_k})$, where $\mathbf{\hat{t}}_k$ is in Equation \ref{eq:pred_per_primitive_type}.

  \item \textbf{Mean point normal difference}:\\
    $\frac{1}{N} \sum_{i=1}^{N} \arccos \left( |\mathbf{N}_{i, :}^\mathbf{T}\mathbf{\hat{N}}_{i, :}| \right)$.

  \item \textbf{Mean primitive axis difference}:\\
    $\frac{1}{\sum_{k=1}^K \mathbb{1}({\mathbf{t}_{k} = \mathbf{\hat{t}}_k})} \sum_{k=1}^K \mathbb{1}({\mathbf{t}_{k} = \mathbf{\hat{t}}_k}) \arccos \left(  \Theta_{\mathbf{t}_k}(\mathbf{A}_k, \mathbf{\hat{A}}_k) \right)$.
    It is measured only when the predicted type is correct.

  \item \textbf{Mean/Std. $\{\mathbf{S}_k\}$ residual}:\\
    $\frac{1}{K} \sum_{k=1}^K \mathbb{E}_{\mathbf{p} \sim U(\mathbf{S}_k)} \sqrt{D^2_{\mathbf{\hat{t}}_k} (\mathbf{p}, \mathbf{\hat{A}}_k)}$. In contrast to the expression for loss $\mathcal{L}_{\text{res}}$, predicted type $\mathbf{\hat{t}}_k$ is used. The $\{\mathbf{S}_k\}$ residual standard deviation is defined accordingly.

  \item \textbf{$\{\mathbf{S}_k\}$ coverage}:\\
    $\frac{1}{K} \sum_{k=1}^K \mathbb{E}_{\mathbf{p} \sim U(\mathbf{S}_k)} \mathbb{1}\left( \sqrt{D^2_{\mathbf{\hat{t}}_k} (\mathbf{p}, \mathbf{\hat{A}}_k)} < \epsilon \right)$, where $\epsilon$ is a threshold.

  \item \textbf{$\mathbf{P}$ coverage}:\\
    $\frac{1}{N} \sum_{i=1}^N \mathbb{1}\left( \min_{k=1}^{K} \left( \sqrt{D^2_{\mathbf{\hat{t}}_k} (\mathbf{P_{i,:}}, \mathbf{\hat{A}}_k)} \right) < \epsilon \right)$, where $\epsilon$ is a threshold.
\end{itemize}
When the predicted primitive numbers is less than $K$, there will be less than $K$ matched pairs in the output of the Hungarian matching. In this case, we modify the metrics of primitive type accuracy, axis difference, and $\{S_k\}$ residual mean/std. to average only over matched pairs.

\subsection{Comparison to Efficient RANSAC~\cite{Schnabel:2007}}
\label{sec:efficient_ransac}
\vspace{-3pt}
We compare the performance of SPFN with Efficient RANSAC~\cite{Schnabel:2007} and also hybrid versions where we bring in predictions from neural networks as RANSAC input. 
We use the CGAL~\cite{CgalRANSAC} implementation of Efficient RANSAC with its default adaptive algorithm parameters.
Following common practice, we run the algorithm multiple times ($3$ in our all experiments), and pick the result with highest input coverage. Different from our pipeline, Efficient RANSAC requires point normals as input. We use the standard jet-fitting algorithm~\cite{Cazals:2003} to estimate the point normals from the input point cloud before feeding to RANSAC.

We report the results of SPFN and Efficient RANSAC in Table~\ref{tbl:results}. Since Efficient RANSAC can afford point clouds of higher resolution, we test it both with the identical $8k$ input point cloud as in SPFN (row 1), and with another $64k$ input point cloud sampled and perturbed in the same way (row 2). Even compared to results from high-resolution point clouds, SPFN outperforms Efficient RANSAC in all metrics. Specifically, both $\{\mathbf{S}_k\}$ and $\mathbf{P}$ coverage numbers with threshold $\epsilon=0.01$ show big margins, demonstrating that our SPFN fits primitives more precisely.

We also test Efficient RANSAC by bringing in per-point properties predicted by SPFN. 
We first train SPFN with only $\mathcal{L}_\text{seg}$ loss, and then for each segment in the predicted membership matrix $\mathbf{\hat{W}}$ we use Efficient RANSAC to predict a single primitive (Table~\ref{tbl:results}, row 4).
We further add $\mathcal{L}_\text{type}$ and $\mathcal{L}_\text{norm}$ losses in training sequentially, and use the predicted primitive types $\mathbf{\hat{t}}$ and point normals $\mathbf{\hat{N}}$ in Efficient RANSAC (row 5-6).
When the input point cloud is first segmented with a neural network, both $\{\mathbf{S}_k\}$ and $\mathbf{P}$ coverage numbers for Efficient RANSAC increase significantly, yet still lower than SPFN. 
Notice that the point normals and primitive types predicted by a neural network do not improve the $\{\mathbf{S}_k\}$ and $\mathbf{P}$ coverage in RANSAC. 

\begin{figure}[t]
  \centering
  \includegraphics[width=\columnwidth]{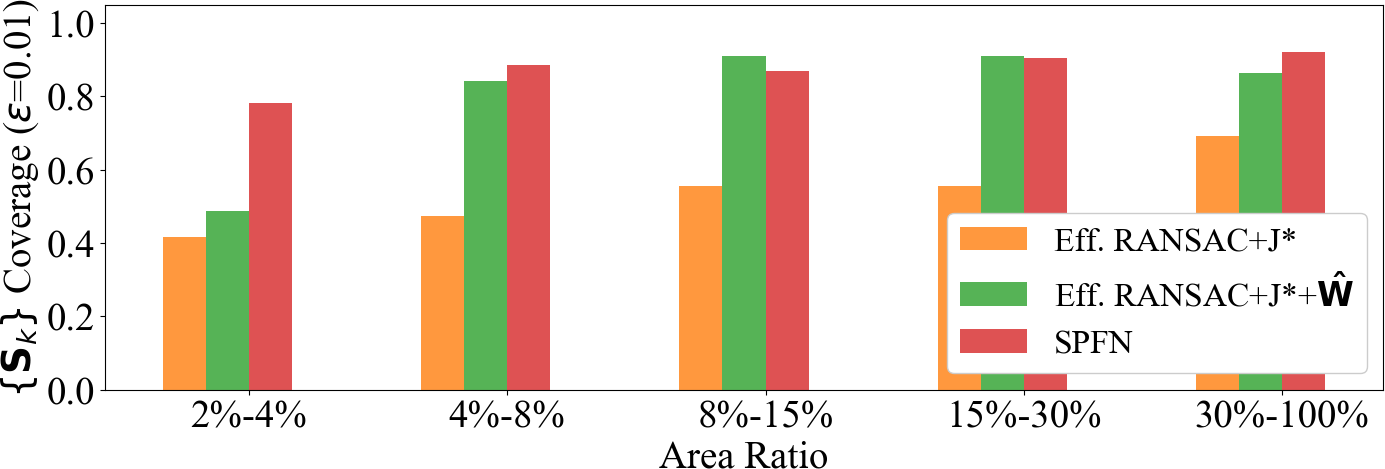}
  \vspace{-1.25\baselineskip}
  \caption{$\{\mathbf{S}_k\}$ coverage against scales of primitives.}
  \label{fig:S_k_coverage}
\vspace{-1.25\baselineskip}
\end{figure}

Figure~\ref{fig:S_k_coverage} illustrates $\{\mathbf{S}_k\}$ coverage with $\epsilon=0.01$ for varying scales of ground truth primitives. Efficient RANSAC coverage improves when leveraging the segmentation results of the network, but still remains low when the scale is small.
In contrast, SPFN exhibits consistent high coverage for all scales.

\subsection{Comparison to Direct Parameter Prediction Network (DPPN)}
\label{sec:dppn}
\vspace{-3pt}
We also consider a simple neural network named Direct Parameter Prediction Network (DPPN) that directly predicts primitive parameters without predicting point properties as an intermediate step. DPPN uses the same PointNet++~\cite{Qi:2017b} architecture that consumes $\mathbf{P}$, \rev{but different from SPFN, it outputs} $K_\text{max}$ primitive parameters for \emph{every} primitive type (so it gives $4K_\text{max}$ sets of parameters). 
In training, the Hungarian matching to the ground truth primitives (Section~\ref{sec:primitives_reordering}) is performed with fitting residuals as in Equation \ref{eq:residual_loss} instead of RIoU. Since point properties are not predicted and the matching is based solely on fitting residuals (so the primitive type might mismatch), only $\mathcal{L}_{\text{res}}$ is used as the loss function. At test time, we assign each input point to the closest predicted primitive to form $\mathbf{\hat{W}}$.

The results are reported in row 7 of Table~\ref{tbl:results}. Compared to SPFN, both $\{\mathbf{S}_k\}$ and $\mathbf{P}$ coverage numbers are far lower, particularly when the threshold is small ($\epsilon=0.01$). This implies that supervising a network not only with ground truth primitives but also with \emph{point-to-primitive} associations is crucial for more accurate predictions.

\subsection{Ablation Study}
\label{sec:ablation_study}
\vspace{-5pt}
We conduct ablation study to verify the effect of each loss term. In Table~\ref{tbl:results} rows 8-11, we report the results when we exclude $\mathcal{L}_{\text{seg}}$, $\mathcal{L}_{\text{norm}}$ (use jet-fitting normals computed from $64k$ points), $\mathcal{L}_{\text{res}}$, and $\mathcal{L}_{\text{axis}}$, respectively. The coverage numbers drop the most when the segmentation loss $\mathcal{L}_{\text{seg}}$ is not used (-$\mathcal{L}_{\text{seg}}$). When using point normals computed from $64k$ input point clouds instead of predicting them (-$\mathcal{L}_{\text{norm}}$+J*), the coverage also drops despite more accurate point normals. This implies that SPFN predicts point normals in a way to better fit primitives rather than to just accurately predict the normals. Without including the fitting residual loss (-$\mathcal{L}_{\text{res}}$), we see a drop in coverage and segmentation accuracy. Excluding the primitive axis loss $\mathcal{L}_{\text{axis}}$ not only hurts the axis accuracy, but also gives lower coverage numbers (especially $\{\mathbf{S}_k\}$ coverage). Row 12 ($\mathbf{\hat{t}} \rightarrow$ Est.) shows results when using predicted types $\mathbf{\hat{t}}$ in the fitting residual loss (Equation $\ref{eq:residual_loss}$) instead of the ground truth types $\mathbf{t}$. 
The results are compatible but slightly worse than SPFN where we decouple type and other predictions in training.

\begin{figure}[t]
  \centering
  \includegraphics[width=0.8\columnwidth]{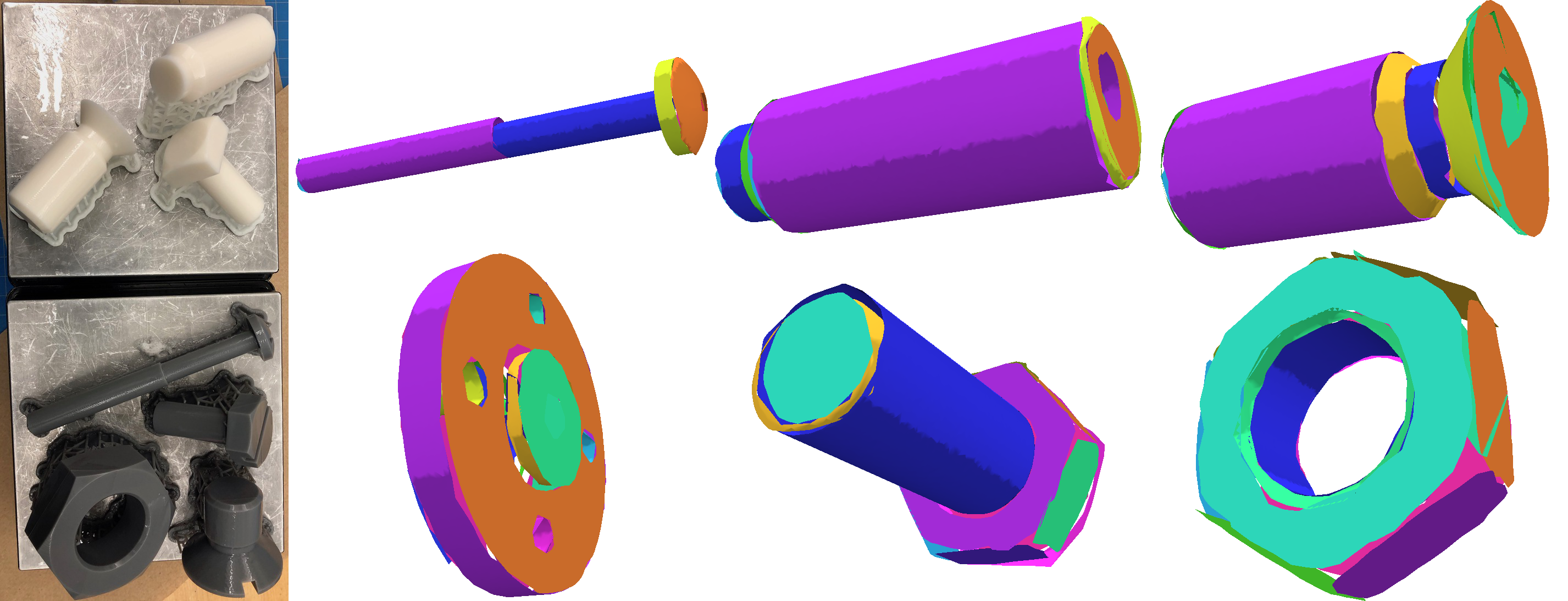}
  \caption{\rev{Results with real scans. Left are the 3D-printed CAD models from the test set.}}
  \label{fig:real_scans}
\vspace{-\baselineskip}
\end{figure}

\subsection{\rev{Results with Real Scans}}
\label{sec:real_scan}
\vspace{-3pt}
\rev{For testing with real noise patterns, we 3D-printed some test models and scanned the outputs using a DAVID SLS-2 3D Scanner. Notice that SPFN trained on synthesized noises successfully reconstructed all primitives including the small segments (Figure~\ref{fig:real_scans}).}

\section{Conclusion}
\label{sec:conclusion}
\vspace{-3pt}
We have presented Supervised Primitive Fitting Network (SPFN), a fully differentiable network architecture that predicts a varying number of geometric primitives from a 3D point cloud, potentially with noise.
In contrast to directly predicting primitive parameters, SPFN predicts per-point properties and then derive the primitive parameters using a novel differentiable model estimator. 
The strong supervision we provide allows SPFN to accurately predict primitives of different scales that closely abstract the underlying geometric shape surface, without any user control. 
We demonstrated in experiments that this approach gives significant better results compared to both the RANSAC-based method~\cite{Schnabel:2007} and direct parameters prediction. 
We also introduced a new CAD model dataset, ANSI mechanical component dataset, along with a set of comprehensive evaluation metrics, based on which we performed our comparison and ablation studies.



\paragraph{Acknowledgments.}
\vspace{-3pt}
The authors wish to thank Chengtao Wen and Mohsen Rezayat for valuable discussions and for making relevant data available to the project. Also, the authors thank TraceParts for providing ANSI Mechanical Component CAD models.
This project is supported by a grant from the Siemens Corporation, NSF grant CHS-1528025 a Vannevar Bush Faculty Fellowship, and gifts from and Adobe and Autodesk. A. Dubrovina acknowledges the support in part by The Eric and Wendy Schmidt Postdoctoral Grant for Women in Mathematical and Computing Sciences.

{\small
\bibliographystyle{ieee}
\bibliography{paper}
}

\clearpage

\renewcommand{\thesection}{S}
\setcounter{table}{0}
\renewcommand{\thetable}{S\arabic{table}}
\setcounter{figure}{0}
\renewcommand{\thefigure}{S\arabic{figure}}

\newif\ifpaper
\papertrue

\section*{Supplementary Material}

\ifpaper
  \newcommand\refpaper[1]{\unskip}
\else
  \makeatletter
  \newcommand{\manuallabel}[2]{\def\@currentlabel{#2}\label{#1}}
  \makeatother
  \manuallabel{fig:results}{4}
  \manuallabel{tbl:results}{1}
  \manuallabel{sec:loss_function}{3.3}
  \manuallabel{eq:sphere_weighted_center}{7}
  \manuallabel{sec:dataset}{4.1}
  \manuallabel{sec:evaluation_metrics}{4.2}
  \manuallabel{tbl:results}{1}

  \newcommand{\refpaper}[1]{in the paper}
\fi

\subsection{Numerical Stability Control}
\label{sec:stability}
In our differentiable model estimator, we are solving two linear algebra problems: homogeneous least square and unconstrained least square. In both problems, numerical stability issues can occur.

We solve the homogeneous least square using SVD to find $\mathbf{v}$, the right singular vector corresponding to the smallest singular value. However, when backpropagating the gradient through SVD~\cite{Ionescu:2015a,Ionescu:2015b}, the gradient value goes to infinity when the singular values of the input matrix are not all distinct.
In our case, such an issue happens only when the output segment (decided by membership matrix $\mathbf{\hat{W}}$) becomes degenerate. For instance, when fitting a plane to a segment via SVD, non-distinct singular values correspond to the case where the points in the segment with significant weights concentrate on a \emph{line} or a \emph{single point}. Hence if we get good segmentation by minimizing the segmentation loss (Section~\ref{sec:loss_function} \refpaper{}), then such degenerate cases should not happen.
Thus, we handle the issue by simply bounding the gradient in the following way. We implemented a custom SVD layer following~\cite{Ionescu:2015a,Ionescu:2015b}, and when computing $K_{ij} = \frac{1}{\sigma_i - \sigma_j}$ in Equation 13 of \cite{Ionescu:2015a} where $\sigma_i$, $\sigma_j$ are singular values, we instead use $K_{ij} = \frac{1}{\text{sign}(\sigma_i - \sigma_j) \max(|\sigma_i - \sigma_j|, \epsilon)}$ for $\epsilon = 10^{-10}$.

When solving the unconstrained least square using Cholesky factorization, numerical unstability can happen even when the segmentation is correct, but the type used in the estimator does not match with the segment. For instance, when fitting a sphere to a segment that is almost a \emph{flat plane}, the optimal sphere is the one with center at infinity. 
To deal with such a singular case (as well as cases when the segments are degenerate), we add a $l_2$-regularizer to the formulation (Equation \ref{eq:sphere_weighted_center} \refpaper{}) and solve instead 
\begin{equation}
\min_{\mathbf{c} \in \mathbb{R}^3} \| \text{diag}(\mathbf w) (\mathbf{X}\mathbf{c} - \mathbf{y})\|^2 + \lambda\|\mathbf{c}\|^2, 
\end{equation}
with $\lambda = 10^{-8}$.
Even with such a modification, Cholesky factorization can still become unstable when the condition number of $\text{diag}(\mathbf{w})\mathbf{X}$ is too large, where the condition number of a matrix is defined to be the ratio of its largest singular value over its smallest singular value.
To deal with this, we trivialize the least square problem when the condition number is larger than $10^5$ by setting $\mathbf{X} = \mathbf{0}$ to prevent gradient flow.

\subsection{Training Details}
\label{sec:training_details}
We use the default hyperparameters for training PointNet++~\cite{Qi:2017b} with a batch size of $16$, initial learning rate $10^{-3}$, and staircase learning decay $0.7$. All neural network models in the experiments are trained for $100$ epochs, using Adam optimizer. The longest experiment (SPFN and its ablation studies) took $50$ hours to train on a single Titan Xp GPU, although the decay of the total loss was not substantial after $50$ epochs.
We will release our source code and include a link to the code in the final version.

\subsection{DPPN Architecture}
\label{sec:DPPN}

\begin{figure}[h]
\vspace{-\baselineskip}
\centering
  \includegraphics[width=\columnwidth]{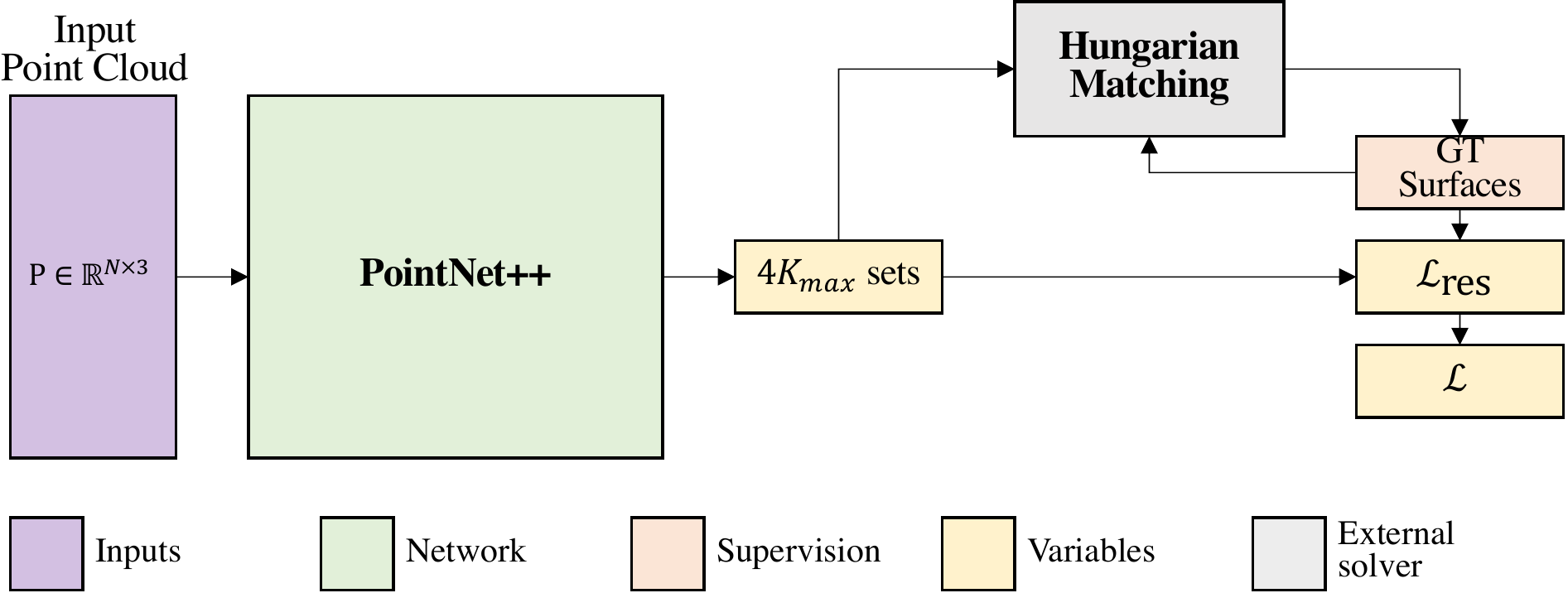}
\vspace{-\baselineskip}
\caption{DPPN architecture.}
\label{fig:dppn_architecture}
\end{figure}
The output of DPPN is simply a collection of $4K_\text{max}$ primitives including $K_\text{max}$ planes, $K_\text{max}$ spheres, $K_\text{max}$ cylinders, and $K_\text{max}$ cones.
In order to compare with SPFN outputs, as a post-processing step, we construct auxiliary membership matrix $\mathbf{\hat{W}}$ by assigning each input point to the closest primitive among the $4K_\text{max}$ predicted primitives.
Similarly, we construct per-point type matrix $\mathbf{\hat{T}}$ by assigning the type of each point to be the type of its closest primitive. The numbers reported in Table~\ref{tbl:results} \refpaper{} are computed in the same evaluation pipeline as in SPFN after such post-processing step.


\begin{table*}[t!]
\centering
\newcolumntype{Y}{>{\centering\arraybackslash}X}
\footnotesize{
{
\setlength{\tabcolsep}{0.2em}
\renewcommand{\arraystretch}{0.9}
\begin{tabularx}{\textwidth}{l|m{3.1cm}|Y|Y|Y|Y|>{\centering}m{1.8cm}|Y|Y|Y|Y}
  \toprule
    \multirow{2}{*}{\scriptsize{Ind}} &
    \multirow{2}{*}{Method} &
    \multirow{2}{*}{{\makecell{Seg.\\(Mean IoU)}}} &
    \multirow{2}{*}{{\makecell{Primitive\\Type (\%)}}} &
    \multirow{2}{*}{{\makecell{Point\\Normal ($^{\circ}$)}}} &
    \multirow{2}{*}{{\makecell{Primitive\\Axis ($^{\circ}$)}}} &
    \multirow{2}{*}{{\makecell{$\{\mathbf{S}_k\}$ Residual\\Mean $\pm$ Std.}}} &
    \multicolumn{2}{c|}{$\{\mathbf{S}_k\}$ Coverage} &
    \multicolumn{2}{c}{$\mathbf{P}$  Coverage} \\
  \cline{8-11}
     & & & & & & &
     $\epsilon = 0.01$ & $\epsilon = 0.02$ & $\epsilon = 0.01$ & $\epsilon = 0.02$ \\
  \midrule
  1 & SPFN (Row 13 in Table 1) &
  77.14 & 96.93 & 8.66 & 1.51 & 0.011 $\pm$ 0.131 & 86.63 & 91.64 & 88.31 & 96.30  \\
  \midrule
  2 & SPFN, $64k$ test input &
  \textbf{77.29} & \textbf{97.27} & \textbf{8.50} & 1.49 & \textbf{0.010 $\pm$ 0.126} & \textbf{87.03} &  \textbf{91.87}  & \textbf{89.01}  &  \textbf{96.42}  \\
  3 & SPFN, w/ outliers &
  72.38 & 95.94 & 9.67 & 1.97 & 0.015 $\pm$ 0.147 & 82.57 &  88.78  & 79.75  &  88.44  \\
  4 & SPFN, $K_\text{max}=48$ &
  76.30 & 96.55 & 8.69 & \textbf{1.39} & 0.011 $\pm$ 0.134 & 85.77 &  90.52  & 88.09  &  95.42  \\
  \bottomrule
\end{tabularx}
}
}
\vspace{-0.5\baselineskip}
\caption{Results of additional experiments described in Section~\ref{sec:additional_experiments}. First row is the same as row 13 in Table \ref{tbl:results} \refpaper{}.
See Section~\ref{sec:evaluation_metrics} \refpaper{} for the details of evaluation metrics. Lower is better in 3-5\textsuperscript{th} metrics, and higher is better in the rest.}
\label{tbl:supp_results}
\end{table*}


\subsection{Primitive Correspondences}
\label{sec:correspondences}

\begin{figure}[!h]
\vspace{-\baselineskip}
  \centering
  \includegraphics[width=\columnwidth]{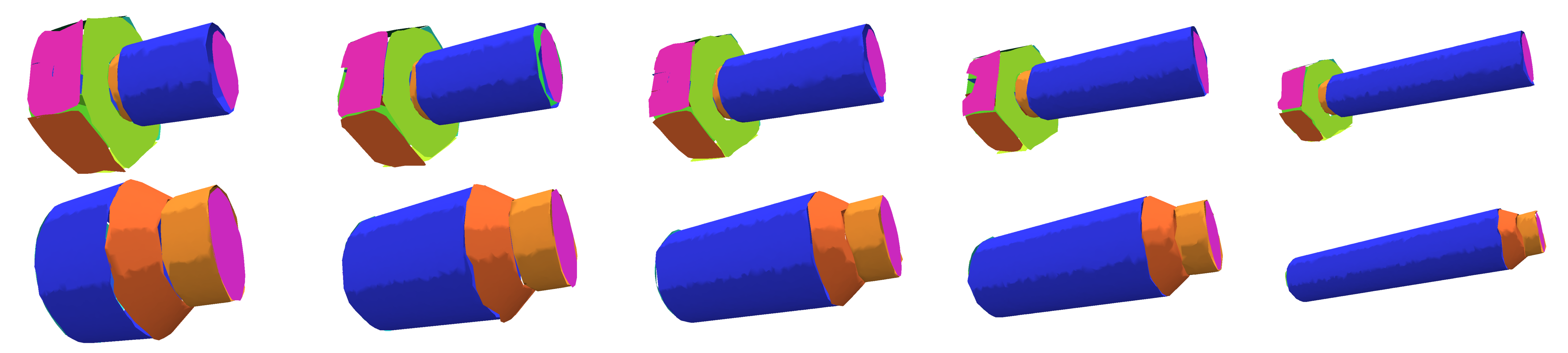}
  \vspace{-\baselineskip}
  \caption{Output primitives of SPFN. Colors indicating column indices in $\mathbf{\hat{W}}$ are consistent across shapes in the same category.}
  \label{fig:correspondences}
\end{figure}

Tulsiani~\etal~\cite{Tulsiani:2017} and Sung~\etal~\cite{Sung:2018} introduced a type of neural networks capable of discovering correspondences across different inputs without direct supervision. Notice in SPFN, changing the ordering of the columns in $\mathbf{\hat{W}}$ does not affect the loss. Despite such ambiguity, SPFN implicitly learns a preferred order such that the primitives represented by the same columns in $\mathbf{\hat{W}}$ in different shapes appear to be similar, resulting in rich correspondence information for primitives from different shapes (Figure~\ref{fig:correspondences}).
These results provide insight into the possible design variations for the same category of shapes.

\subsection{Additional Experiments}
\label{sec:additional_experiments}
\rev{To further study the capability of SPFN, we have conducted the following additional experiments.}

PointNet++~\cite{Qi:2017b} used in our architecture has a limitation of handling high resolution point clouds during training time due to the increase of memory consumption. However, it is also known that PointNet++ is robust to the change of the resolution of point clouds at test time (See Section 3.3 in~\cite{Qi:2017b}). Hence, we can consider processing high resolution input point clouds in the test time by training the network with lower resolution point clouds. In Table~\ref{tbl:supp_results}, row 2 shows the results of testing $64k$ point clouds with the same SPFN model in Table~\ref{tbl:results} \refpaper{} (trained with $8k$ point clouds), and it exhibits a slight improvement in nearly all metrics. We also assessed SPFN by adding not only noise in the inputs (as described in Section~\ref{sec:dataset} \refpaper{}) but also outliers. Row 3 describes the results when we add $10\%$ outliers, which are uniformly sampled in space outside of the central cube $[-0.5,0.5]^3$, in both training and test data. The results show a little drop but still comparable performance. Lastly, we also investigated how robust SPFN can be if we change the maximum number of primitives, $K_\text{max}$. Row 4 illustrated the results when training the network with $K_\text{max}=48$, and we observed no substantial difference in performance.

\end{document}